\pdfoutput=1
\documentclass[11pt]{article}
\usepackage{times}
\usepackage{graphicx} % Required for inserting images
\usepackage{subcaption}
\usepackage{ACL2023}
\usepackage[T1]{fontenc}
\usepackage[utf8]{inputenc}
\usepackage{microtype}
\usepackage{inconsolata}
\usepackage{tabularx}
\usepackage{array}
\usepackage{comment}
\usepackage{booktabs}
\usepackage{amsmath}
\usepackage{makecell}

\title{Pointer-Generator Networks for Low-Resource Machine Translation: Don't Copy That!}

\author{Niyati Bafna, Philipp Koehn, and David Yarowsky\\
         Johns Hopkins University, Center for Language and Speech Processing  \\
         \{nbafna1,phi,yarowsky\}@jhu.edu}

\begin{document}
\maketitle
\begin{abstract}
While Transformer-based neural machine translation (NMT) is very effective in high-resource settings, many languages lack the necessary large parallel corpora to benefit from it. In the context of low-resource (LR) MT between two closely-related languages, a natural intuition is to seek benefits from structural ``shortcuts’’, such as copying subwords from the source to the target, given that such language pairs often share a considerable number of identical words, cognates, and borrowings. We test Pointer-Generator Networks for this purpose for six language pairs over a variety of resource ranges, and find weak improvements for most settings ($<1$ BLEU). However, analysis shows that PGNs do not show greater improvements for closely-related vs. more distant language pairs, or for lower resource ranges, and that the models do not exhibit the expected usage of the mechanism for shared subwords. Our discussion of the reasons for this behaviour highlights several general challenges for LR NMT, such as modern tokenization strategies, noisy real-world conditions, and linguistic complexities. We call for better scrutiny of linguistically motivated improvements to NMT given the blackbox nature of Transformer models, as well as for a focus on the above problems in the field.
\end{abstract}

\section{Introduction and Motivation}
\label{sec:intro}
% NMT does well for HRL pairs
While state-of-the-art (SOTA) Transformer models \citep{vaswani2017attention} for NMT work well for high-resource language pairs, their performance degrades in low-resource situations \citep{koehn-knowles-2017-six,sennrich-zhang-2019-revisiting,kim-etal-2020-unsupervised,haddow-etal-2022-survey}; this means that most languages in the world cannot benefit from mainstream advances and models \citep{joshi-etal-2020-state}. There is therefore a clear appeal to developing simple architectural mechanisms for these models that are targeted at yielding improvements in data-scarce scenarios, while interfering minimally with mainstream preprocessing, tokenization, and training pipelines.

\begin{figure}
    \centering
    \includegraphics[scale=0.22]{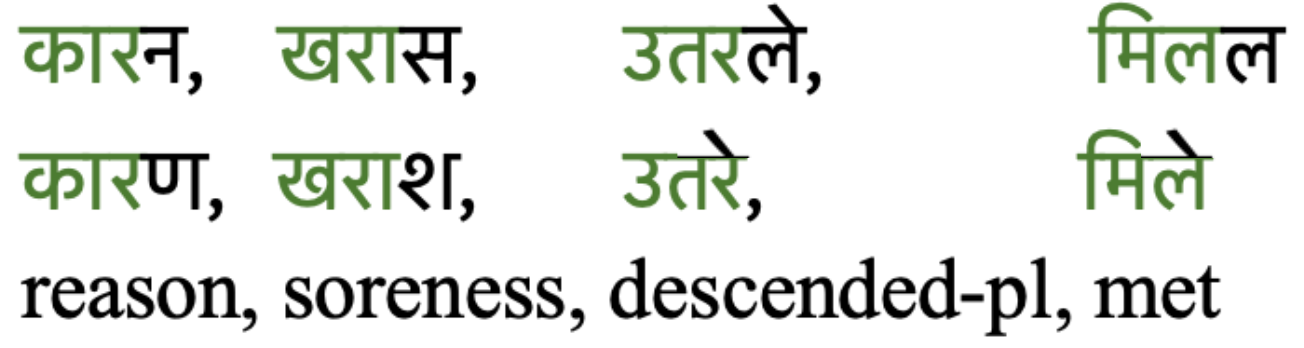}
    \caption{Translation equivalents for Bhojpuri (top) and Hindi (bottom), demonstrating subword overlap.}
    \addtolength{\belowcaptionskip}{-100pt}
    \label{fig:cognates}
\end{figure}

% Many LRLs in the world - we frequently want to translate to and from a closely related, often high-resource language
In the context of a low-resource language (LRL), we are often interested in translation to and from a closely related HRL, which possibly has linguistic genealogical, regional, and cultural ties with the LRL,\footnote{This is the case, for example, for several languages of the Arabic continuum, all closely related to relatively high-resource Modern Standard Arabic, and languages of the Turkic and Indic language continua.}  in order to make the abundant content in HRLs available in related LRLs.
%The problem of translating between related languages is highly relevant to making the abundant content in the world's HRLs available to related LRLs.This can even be via pivot translation e.g. we may have very good MT for English-Hindi but not English-Bhojpuri; in this case, it would be helpful to have good quality Hindi-Bhojpuri systems.}
% These languages may share cognates, borrowings, and cultural references.
We expect that closely related languages share considerable overlap at the subword level from cognates, borrowings and shared vocabulary (see examples in Figure~\ref{fig:cognates}). Given the absence of large parallel corpora for our language pair, we aim to leverage this shared knowledge across source and target, intuitively, to provide ``easier'' routes for our MT model from source to target sentence. 

% PGNs: a mechanism by which one can optionally copy from source 
Pointer Generator Networks (PGNs; \citet{see2017get}) are a mechanism which allow the model, for every output token produced, to either copy some token from the input (``point'') or ``generate'' a token as per usual from the vocabulary. PGNs have been used for a variety of problems, described in Section~\ref{sec:rel_work}, often targeted at repeated spans of text in the input and output; however, as we far as we know, this is the first work to study its applicability to LR NMT. In this case, we hypothesise that the pointing mechanism will show advantages for rare shared subwords, for which the best strategy may be to copy them to the output.

We introduce a PGN mechanism into a Transformer-based NMT architecture, and test its performance for 6 language pairs over 4 low-resource training ranges. We work with Hindi-Bhojpuri (hi-bh), Spanish-Catalan (es-ca), and French-Occitan (fr-oc), representing closely-related pairs, Hindi-Marathi (hi-mr), a relatively more distant pair,\footnote{Hindi, Bhojpuri, and Marathi belong to the Indic branch of the Indo-European family. Hindi and Bhojpuri further belong to the Shaurasenic sub-branch and are closer lexically and grammatically to each other and other languages on or close to the Hindi Belt such as Punjabi, Rajasthani, Haryanvi, and Maithili, than Hindi is to Marathic languages and dialects; this is supported by lexical and other studies of cross-lingual similarity \citep{sengupta2015study,mundotiya2021,bafna-etal-2022-combining}. See Glottolog (\url{https://glottolog.org/resource/languoid/id/cont1248}) for the phylogenetic tree.} and Spanish-English (es-en) and French-German (fr-de), representing further distant pairs. We expect that \textsc{PGN} will help most for (1) lower-resource scenarios (2) more closely-related language pairs (3) sentence pairs with higher subword overlap. While \textsc{PGN} shows improvements in certain settings, our comparative analysis of the benefits of PGNs across the three dimensions above shows clear lack of evidence for the these hypotheses. Further, our visualizations of the PGN mechanism also indicate that observed benefits do not come from intended sources. We discuss various factors that contribute to this failure, highlighting fundamental challenges for LR NMT, such as noisy datasets, mainstream tokenization practices best suited for high-resource scenarios, as well as linguistic and orthographic complexities that may obfuscate underlying source-target similarities.\footnote{\url{https://github.com/niyatibafna/pgns-for-lrmt}}
%\footnote{Our code is available at \url{https://anonymous.4open.science/r/pgns-for-lrmt-38DE}.} 

%We also underline the need for better analysis of linguistic motivated improvements in NMT, given the unknown generalization mechanism of these architectures that are often not amenable to linguistic intuition, and call for a focus on these fundamental challenges in the field.

 \section{Related Work}
\label{sec:rel_work}

Pointer Networks were introduced to solve problems that involved permuting the input, such as the Traveling Salesman Problem and the complex hull problem \citep{vinyals2015pointer}. Their use in NLP has been largely been for monolingual summarization, where the target may naturally contain identical spans from the source. \citet{cheng-lapata-2016-neural} present a complex hierarchical LSTM-based model for summarization, which directly extracts sentences from the text and words from sentences. \citet{gulcehre2016pointing} use pointer networks in RNN-based sequence-to-sequence models for summarization and machine translation, training their model explicitly to use the pointing mechanism for uncommon words. \citet{gu2016incorporating} and \citet{see2017get} also incorporate variants of pointer-generator networks into RNN-based sequence-to-sequence learning for summarization. 
\citet{prabhu2020making} applied PGNs to the task of grapheme-to-phoneme conversion via an explicit source-target mapping. \citet{zhang2021point} proposed a pointer-disambiguator-copier (PDC) system for dictionary-enhanced high-resource NMT, using source word translations as potential candidates for the copying mechanism, with a disambiguator component to select appropriate senses.

Our work is the first to examine the applicability of PGNs as facilitators in the low-resource MT scenario, looking to exploit linguistic relationships between the source and target in the absence of external resources. We work with Transformer-based NMT, and make no changes to standard BPE tokenization schemes or training objectives (unlike \citet{gulcehre2016pointing} and \citet{zhang2021point}). This is so that our findings are most relevant in today's paradigm of generalized strategies for end-to-end multilingual MT; our mechanism can be easily plugged into and trained with any modern (multilingual) MT pipeline.

\section{Model}
\label{sec:model}
The \textsc{PGN} model provides two routes to the model for predicting any target token: copying from the source or generating from the vocabulary \citep{prabhu2020making}. Copy and generate distributions at step $t$ are mixed using a learned parameter $p_{copy}^{t}$, to obtain the final probability distribution $P^t$ for the target token. 
\begin{align}
&p_{copy}^{t}=\sigma(\textbf{W}^T(\textbf{c}^{t} \oplus \textbf{d}^{t} \oplus \textbf{s}^{t}) + \textbf{B}) \nonumber\\
&\textbf{P}^{t} = p_{copy}^{t} \cdot \textbf{P}_{\textbf{c}}^{t} + (1 - p_{copy}^{t}) \cdot \textbf{P}_{\textbf{g}}^{t} \nonumber
\end{align}

Here, $ \textbf{c}^{t} $ is the context vector, calculated as $ \textbf{c}^{t} = (\textbf{a}^{t})^T \textbf{e}^{t} $, where $ \textbf{a}^t $ represents cross-attention vector, and $ \textbf{e}^t $ contains the encoder hidden states. $ \textbf{d}^{t} $ and $ \textbf{s}^{t} $ contain the decoder's final hidden states and input respectively, $\oplus$ denotes concatenation, and $ \textbf{W} $ and $ \textbf{B} $ are learned weights and a bias vector respectively.  
$\textbf{P}_{\textbf{c}}^{t}$ and $\textbf{P}_{\textbf{g}}^{t}$ represent the copy and generate distributions (softmaxed logits) respectively at step $t$. We use cross-attention weights over source tokens for the copy logits, and standard decoder outputs for generate logits.

%In practice, this is achieved by the following method: The cross-attention weights are obtained by averaging over heads, and a context vector is found by taking a weighted sum of the encoder hidden states. The context vector is then concatenated with the decoder hidden states and the decoder input, and the resulting vector is passed through a learned linear layer. 
 %In practice, this is done by obtaining generate logits from the decoder outputs and copy logits from the cross-attention weights, then normalizing these to obtain generate and copy probabilities from the decoder. We then mix our various probabilities by multiplying the generate probability by $p_{gen}$ and copy probability by $p_{copy}$, then adding the two results to obtain $P(k)$. After this, we calculate negative log likelihood loss function against the true targets.

\section{Experiments}

\paragraph{Datasets and languages}
\label{sec:data}

We used the WikiMatrix (\texttt{wm)} corpus \citep{schwenk2019wikimatrix} for \texttt{es-en}, \texttt{es-ca}, \texttt{fr-de} and \texttt{fr-oc}. For \texttt{es-ca}, we also report results on synthetic Europarl (\texttt{ep}) parallel data \citep{koehn-2005-europarl}.\footnote{The Europarl dataset was automatically translated into Catalan; taken from \url{https://github.com/Softcatala/Europarl-catalan}.} For \texttt{hi-mr}, we used the CVIT-PIB corpus \citep{Philip_Siripragada_Namboodiri_Jawahar_2021}, and for the low-resource pair \texttt{hi-bh}, we use the NLLB corpus \citep{schwenk-etal-2021-ccmatrix,nllbteam2022language,heffernan2022bitext}. See Table~\ref{tab:tokens} for dataset heuristics. Note the higher per token overlap as expected for our closely-related group as compared to the others. The \texttt{hi-bh} sentences are extremely short, and share fewer tokens than expected: in this case, this reflects badly parallel data.\footnote{See Appendix~\ref{app:langs} for more details on datasets.}
%The \texttt{hi-mr} and \texttt{hi-bh} datasets are both highly noisy, the latter more so; these represent difficult conditions for MT. Meanwhile, the \texttt{es-ca(wm)}, \texttt{es-ca(ep)}, and \texttt{fr-oc} data are much cleaner and better aligned.\footnote{See App~\ref{app:langs} for more details on datasets.}

% Notes:
% Hindi-Marathi WikiMatrix has only 11k sentences so we use CVIT-PIB instead.
% Spanish-Catalan Europarl : Catalan Europarl is translated using Apertium (cite) so it exhibits translationese. It's more likely to contain literal, linear translations, and represents an easier scenario for our experiments.
% We looked at other Hindi-related languages, such as Bhojpuri and Magahi. NLLB datasets for these pairs (including Hindi-Marathi) are extremely noisy and badly parallel. This underscores our point about real-world conditions...

% \begin{table*}[t]
%     \centering
%     \begin{tabular}{lrr}
%     \hline
%     {} & \textbf{Es-Ca} & \textbf{Hi-Mr}\\
%     \hline
%     % \textbf{Common tokens} & {12702359} & {173784} \\
%     % \textbf{Total target tokens} & {49212670} & {1441496} \\
%     \textbf{\#Sentences} & {1876669} & {69186} \\
%     \textbf{Avg. common/sentence} & {6.77} & {2.51} \\
%     \textbf{Avg. tokens/tgt token} & {0.26} & {0.12} \\
%     \hline
%     \end{tabular}
%     \caption{Statistics on commons tokens for Spanish-Catalan and Hindi-Marathi}
%     \label{tab:tokens}
%     %Average number of tokens in Spanish 26.5 in Hindi 29.6
% \end{table*}

\begin{table*}[htbp]
\small
    \centering
    \begin{tabular}{lccccccc}
        \toprule
        & \texttt{hi-mr} & \texttt{hi-bh} & \texttt{es-en} & \texttt{es-ca (ep)} & \texttt{es-ca (wm)} & \texttt{fr-de} & \texttt{fr-oc} \\
        \midrule
%        \textbf{Common Tokens} & 157820 & 77521 & 338505 & 430135 & 303327 \\
%        \textbf{Total Target Tokens} & 1311475 & 478769 & 1313162 & 1478520 & 1405512 \\
%        \textbf{Total Lines} & 63000 & 60000 & 50000 & 60000 & 60000 \\
        Avg. common tokens per sent pair & 2.51 & 1.29 & 2.81 & 6.77 & 7.17 & 1.76 & 5.06 \\
        Avg. common tokens per target token & 0.12 & 0.16 & 0.13 & 0.26 & 0.29 & 0.10 & 0.22 \\
        Avg. source sentence length & 28.54 & 6.34 & 23.43 & 26.86 & 25.38 & 19.10 & 24.09 \\
        Avg. target sentence length & 20.82 & 7.98 & 20.77 & 26.26 & 24.64 & 16.93 & 23.43 \\
        \bottomrule
    \end{tabular}
    \caption{Statistics on commons source-target tokens in our datasets. \texttt{wm}: WikiMatrix, \texttt{ep}: Europarl. }
    \label{tab:tokens}
\end{table*}

\paragraph{Experimental Settings}

For all language pairs, we performed experiments on dataset subsets of $5k$, $15k$, $30k$, and $60k$ sentences and test sets of $5000$ sentences, trained until convergence, with tokenizer size $16000$. We computed baseline results (\textsc{NMT}) on standard encoder-decoder NMT. All \textsc{PGN} and \textsc{NMT} models use $6$ encoder and decoder layers, $4$ attention heads, and a hidden size of $512$.

\section{Results and Discussion}
\label{sec:disc}
\paragraph{Improvement patterns}
\begin{table*}[ht]
\tiny
    \centering
    \begin{tabularx}{\textwidth}{l*{6}{>{\raggedleft\arraybackslash}X}}
        \toprule
        & &  5K & 15K  & 30K  & 60K  & Avg. $\Delta$\\
        \midrule
        % & \multicolumn{4}{c}{hi-mr} \\
        % \midrule
        \texttt{hi-mr} & \textsc{NMT} & 3.4 & 7.4 & 11.9 & 16.3 & \\
        & \textsc{PGN} & 3.4 & \textbf{7.9} & \textbf{12.6} & \textbf{18.4} & +0.8 \\
        \midrule
        % & \multicolumn{4}{c}{hi-bh} \\
        % \midrule
        \underline{\texttt{hi-bh}} & \textsc{NMT} & \textbf{5.6} & 6.1 & 9.7 & \textbf{15.8} & \\
        & \textsc{PGN} & 4.0 & \textbf{8.3} & \textbf{11.4} & 12.7 & -0.2\\
        \midrule
        \texttt{es-en} & \textsc{NMT} & \textbf{9.8} & \textbf{30.4} & 38.3 & 41.7 & \\ 
        & \textsc{PGN} & 9.4 & 30.0 & \textbf{38.5} & \textbf{42.2} & 0.0 \\
        \midrule
        % & \multicolumn{4}{c}{es-ca,wm} \\
        % \midrule
        \underline{\texttt{es-ca(wm)}} & \textsc{NMT} & \textbf{35.4} & 50.9 & \textbf{54.3} & 56.4 &  \\
        & \textsc{PGN} & 34.7 & \textbf{51.2} & 54.1 & \textbf{57.1} & 0.0\\
        \midrule
        % & \multicolumn{4}{c}{es-ca}  \\
        % \midrule
        \underline{\texttt{es-ca(ep)}} & \textsc{NMT} & \textbf{62.6} & 70.6  & 73.2 & 73.6 & \\ 
        & \textsc{PGN} & 62.5 &  \textbf{71.6} & \textbf{74.0} & \textbf{74.2} & +0.6 \\ 
        \midrule
        \texttt{fr-de} & \textsc{NMT} & 3.5 & 10.8 & 19.5 & 27.2 & \\
        & \textsc{PGN} & \textbf{3.7} & \textbf{11.1} & \textbf{20.1} & 27.2 & +0.3 \\
        \midrule
        % & \multicolumn{4}{c}{fr-oc} \\
        % \midrule
        \underline{\texttt{fr-oc}} & \textsc{NMT} & 24.7 & 42.2 & 45.5 & \textbf{48.7} & \\
        & \textsc{PGN} & \textbf{24.8} & \textbf{43.4} & \textbf{46.4} & 48.5 & +0.5\\
        \bottomrule
    \end{tabularx}
    \caption{spBLEU across dataset sizes (\#sents). Closely-related pairs are \underline{underlined}. \texttt{wm}: WikiMatrix, \texttt{ep}: Europarl.}
    \label{tab:results}
\end{table*}

\begin{table}[ht]
\centering
\tiny
\begin{tabular}{llccccccc}
\toprule
& & \texttt{hi-mr} & \underline{\texttt{hi-bh}} & \texttt{es-en} & \underline{\texttt{es-ca}} & \texttt{fr-de} & \underline{\texttt{fr-oc}} & \makecell{Avg. \\ $\Delta$} \\ 
\midrule
L &  \textsc{NMT} & 7.6 & 10.4 & \textbf{26.6} & \textbf{44.9} & \textbf{11.9} & \textbf{35.5} & \\ 
& \textsc{PGN} & \textbf{8.2} & \textbf{13.5} & 24.4 & 44.6 & 10.3 & 35.3 & -0.7 \\
\midrule
H & \textsc{NMT} & 28.9 & \textbf{22.1} & 69.8 & \textbf{72.4} & \textbf{52.2} & \textbf{65.2} & \\ 
& \textsc{PGN} & \textbf{29.3} & 18.3 & \textbf{69.9} & 71.6 & 51.6 & 64.3 & -1.0 \\ 
\bottomrule
\end{tabular}
\caption{spBLEU scores on test sets with low (L) and high (H) density of shared source-target subwords.}
\label{tab:low_high}
\end{table}

See results in Table~\ref{tab:results}.\footnote{We report spBLEU since our approach attempts to benefit performance on shared subwords.} Our results are not directly comparable to those in the literature due to differences mentioned in Section~\ref{sec:rel_work} and the size of training bitext ($2M$ in \citep{gulcehre2016pointing}, $~1M$ in \citet{zhang2021point} vs. our maximum resource setting of $0.06M$).\footnote{For a rough idea: \citet{zhang2021point} report gains in MT of $1.5-2.5$ \texttt{BLEU}.} We see weak improvements in a majority of settings; however, counter to intuition, \textsc{PGN} does not show a clear advantage for closely-related as compared to more distant pairs, or for lower-resource settings. 

\paragraph{Controlled test sets} We test the motivating hypothesis that \textsc{PGN} models is likely to benefit sentence pairs with higher subword overlap. We rank sentence pairs in our test set by percentage of shared subwords in source and target, and construct test subsets with low and high shared-subword density from the top and bottom $500$ sentences respectively. However, in Table~\ref{tab:low_high}, we see that in fact that \textsc{PGN} performs slightly worse than \textsc{NMT} on both extremes, indicating that observed benefits over the entire test set do not come from subword overlap. 

\paragraph{Usage of the copy mechanism}
We record values of $p_{copy}$ to track the model's usage of the copy mechanism. While $p_{copy}$ values are relatively high\footnote{Note that it is difficult to comment on absolute values of $p_{copy}$. The copying distribution is normalized over the sentence length whereas generate distributions are normalized over the vocabulary; even low values of $p_{copy}$ will considerably affect the mixed distribution.} for copied subwords, numerals, and proper nouns, we often see that they they are also high for seemingly random subwords.\footnote{See Appendix~\ref{app:vis} for visualizations of this behaviour. Examples with counter-intuitively high values of $p_{copy}$: \texttt{quiero-vull (es-ca), behad-atishay (hi-mr)}.} We also do not see a relationship between the $p_{copy}$ value of a target token and the entropy of the cross-attention distribution for that token. 

A reasonable intuition about \textsc{PGN} training generalization is that in the absence of any information, the model will default to copying, since this is likely to do better on average than guesses over the entire vocabulary, and that eventually, it will learn to generate language-specific subwords, memorising the relevant strategy for given subwords in encoder states (used to calculate $p_{copy}$ as shown in Section~\ref{sec:model}). However, our visualizations of cross-attention and $p_{copy}$ usage throughout training show no evidence of this generalization strategy. It's possible that since initial cross-attention distributions are noisy, and most subwords are not direct copies, the model is discouraged from copying early on; it's also possible that the model finds it easier or trivial to encode copied source-target equivalents via the ``generate'' mechanism and does not need an explicit copier, given that it must additionally learn which subwords should be copied. We discuss potential reasons for this below. In general, it appears that the model uses the copy mechanism to encode a task that is not easily interpretable, possibly resulting in the observed small improvements over some datasets.
%, and highlight the need to verify that observed improvements come from expected theoretical sources. 

\paragraph{Tokenization}
\label{sec:tok}
In theory, the copier would learn best if the tokenizer behaved in a morphologically principled manner.\footnote{e.g. given \texttt{khaya-khalla (ate)} in Hindi and Marathi, we ideally want \texttt{kha \#\#ya} and \texttt{kha \#\#lla}. This will allow the common stem \texttt{kha} to be copied over, while the language specific inflection subwords can be generated.} However, BPE tokenization generally results in subword splits that may not reflect shared stems in word equivalents \citep{ataman2018evaluation}.\footnote{e.g. our trained tokenizer contains both \texttt{propuesto} (\texttt{es}) and \texttt{proposat} (\texttt{ca}) instead of sharing the subword \texttt{prop}.}
%$\footnote{\citet{zhang2021point} overcome this problem using ``selective BPE'' i.e. explicitly leaving certain words untouched in the tokenizer. However, such ideas require a compilation of such words for each language pair, and are unlikely to be widely adopted into modern multilingual tokenization strategies.} 

A natural idea here may be an investigation of morphologically inspired tokenizers \citep{pan2020morphological,ortega2020neural,chen-fazio-2021-morphologically}. However, we generally see inconclusive, at best marginal, benefits of such tokenizers over BPE in modern neural MT \citep{macháček2018morphological,domingo2019does,mielke2021words}, especially those relying on unsupervised morphological segmentation, e.g., with Morfessor \citep{creutz2007unsupervised} in the absence of morphological analysers. These ideas have not been incorporated into mainstream tokenization strategies.

Recent work attempts to solve this general problem by looking at maximisation of shared subwords in multilingual tokenizers \citep{chung-etal-2020-improving,zheng-etal-2021-allocating,liang_2023}; it's possible that such strategies will dovetail well with PGN mechanisms if widely adopted in the future. 

\paragraph{Linguistic complexities}
\label{sec:obs_phen}
While closely-related language show high (subword) vocabulary overlap, word equivalences may be obscured by sound change and orthographic systems; if these changes are word-internal, then even an ideal tokenizer will see different stems/tokens in the source and target.\footnote{e.g. \texttt{\textbf{v}i\textbf{sh}was-\textbf{b}i\textbf{s}was} (\texttt{hi-bh}, sound change), \texttt{website-Webs\textbf{e}ite} (\texttt{en-de}, orthographic system)} Further, we may see that a word that has a cognate in its sister language is translated to a non-cognate,  due to semantic drift,
%\footnote{This refers to the phenomenon by which word senses evolve over time; therefore, cognate words in two different languages may mean different things, or be ``false friends''.} 
or differences in idiom or usage norms in the two languages, e.g. \texttt{kitaab-pustak (hi-mr)}, 
resulting in non-identical subword equivalences. These phenomena are often unpredictable and unsystematic; even if not, they are not trivial to model into tokenization or architectural strategies for MT.
\newline \newline
See Appendix~\ref{app:minor} for experiments with minor variants of our approach dealing with pretrained encoder/decoder initialization, tokenizer size, choice of attention head, and identical source-target settings.

%\nb{A possible thing that may be logical to try: add in the linear layer but over the whole vocabulary - this checks whether gains are simply coming from additional parameters.}
% \section{Subword overlap in closely-related languages}

\begin{comment}
    
\subsection{Datasets and non-literal translations}

Translations may be non-literal - in fact, the better they are, the more this will be the case. While Europarl documents are translated with high precision and exhibit linear word-to-word correspondences, this is an exceptional scenario; we expect that parallel corpora will often contain loose translations, especially if they are mined from the web. This reduces the viability of the copying mechanism. We want our training strategy to be robust to loosely translated bitext...
\end{comment}

\section{Challenges for LR NMT}
\label{sec:challenges}
%MT between regional languages constitutes a relevant focus for NLP. 
Incorporating knowledge of linguistic relationships among closely related data-imbalanced language pairs offers a natural strategy for mitigating data scarcity in mainstream NMT between regional languages, and it is crucial to understand the challenges in this realm. We show that while the PGN mechanism offers an intuitive theoretical shortcut for translation between closely related languages, its performance in practice is limited, potentially by the combined effect of noisy real-world datasets containing non-literal translations, the behaviour of standard tokenizers, as well as linguistic complexities beyond simplified ideas of shared vocabulary and cognates. These are inevitable hurdles to any project that attempts to use structural linguistic knowledge to benefit NMT performance. 

Further, we show that despite showing benefits in certain settings over the entire test set, the \textsc{PGN} mechanism does not perform as expected on target phenomena. The generalization mechanisms of blackbox Transformer-models are not well understood and may not be easily guided by linguistic intuition: we underline the importance of verifying that improvements are coming from the intended places rather than good starts or extra parameters.

Finally, our analysis hints that \textsc{PGN}-like shortcuts may not be worth offering in the first place: ``easy'' equivalences, a natural target of linguistic interventions, may not be the bottleneck for LR NMT. Instead, it is more likely that the true bottlenecks are handling precisely the above challenges, i.e. non-systematic differences, one-off phenomena, and real-world noise in low-resource conditions. 
%We underline the need for quality analysis of linguistic motivated improvements in NMT, , and call for a focus on the above fundamental challenges in the field.
%We believe that there is scope for fundamental progress in dealing with these challenges, and call for better analysis of linguistically motivated approaches…

\section{Limitations}
\label{sec:limitations}
While we show that our particular flavour of NMT incorporating PGNs does not provide fundamental benefits for low-resource NMT, this is naturally not to say that an improved variant of this idea would not work better. There are several potential ways forward arising from our discussion of the reasons for the failure of our method in Section~\ref{sec:disc}: for example, using morphological segmentation for tokenization to increase subword overlap, or using priors for $p_{copy}$ so that it is encouraged for shared subwords. Previous work provides different kinds of help to the copier: for example, \citep{gulcehre2016pointing} explicitly train the copier to copy unknown words with a separate training objective. However, as we mention in Section~\ref{sec:intro}, our motivation lay in designing a simple architectural mechanism that can be easily integrated into mainstream (multilingual) NMT pipelines to make them more capable for low-resource MT, without requiring much additional language-pair specific attention to training paradigms or tools such as morphological analysers and bilingual lexicons, which are in any case of poor quality for low-resource languages. We restrict our negative result to the scope set up by this motivation. 

Further, our results are limited to the $6$ language pairs that we experimented on. While we simulate identical low-resource conditions for all our language pairs, we clearly see the difference in absolute performances on \texttt{hi-mr} or \texttt{hi-bh} as compared to the high-resource language pairs: the data for the latter are simply of much better quality. This demonstrates the need to experiment and present further results on non-simulated truly low-resource conditions, such as the hi-bh language pair studied here. Finally, this discussion is only relevant to translation between closely-related languages that share a script (although this is the predominant case), allowing for lexical similarity to be reflected by shared subwords. 

\section{Conclusion}
In this work, we investigate the applicability of Pointer-Generator Networks in NMT, hypothesizing that an explicit copy mechanism will provide benefits for low-resource translation between closely related languages. We show that while we do observe weak improvements, these are not higher for closer-related languages, sentence pairs with higher overlap, or lower resource ranges, contrary to intuition. Our discussion of potential reasons for the failure of this approach highlights several general challenges for low-resource NMT, such as mainstream tokenization strategies, noisy data, and non-systematic linguistic differences.

\newpage
\bibliography{main}
\bibliographystyle{acl_natbib}

\newpage

\appendix

% \section{Language Pairs}
% \label{app:langs}
%\nb{Discuss closeness, comment on Table~\ref{tab:tokens}.} 
\newpage

\section{Notes on Datasets}
\label{app:langs}
The \textbf{Hindi-Marathi} WikiMatrix dataset \citep{schwenk-etal-2021-ccmatrix} has only 11k sentences so we use CVIT-PIB \citep{Philip_Siripragada_Namboodiri_Jawahar_2021} instead. The CVIT-PIB corpus is automatically aligned using an iterative process that depends on neural machine translation into a pivot language and filtering heuristics. Eyeballing the data, we observe a considerable number of non-parallel or even entirely unrelated sentences simply containing some words in common - in general, this corpus is much more likely to contain rough paraphrases as opposed to literal translations. 

For \textbf{Hindi-Bhojpuri}, NLLB seems to be the only available parallel text data for now \citep{tiedemann-2012-parallel}; this corpus has also been automatically crawled \citep{nllbteam2022language}. The Hindi-Bhojpuri NLLB dataset contains extremely short sentences as shown in Table~\ref{tab:tokens}; similarly to above, we observe a high level of noise and non-parallel data. Such datasets naturally do not provide the most favourable training conditions for the \textsc{PGN} models, which rely on literal translations containing shared subwords to teach the copier; however, they are realistic real-word conditions for truly low-resource languages, as we discuss in Section~\ref{sec:challenges}.

The WikiMatrix dataset \citep{schwenk2019wikimatrix}, which we use for \textbf{Spanish-English}, \textbf{Spanish-Catalan}, \textbf{French-German}, and \textbf{French-Occitan} is automatically aligned from Wikipedia content in these languages.

The \textbf{Spanish-Catalan} synthetic Europarl bitext is created by automatically translating the Europarl dataset \citep{koehn-2005-europarl} into Catalan using Apertium \citep{forcada2011apertium,khanna2021recent}. While this data probably contains some noise due to MT errors and translationese, it's the most likely of all our datasets to contain literal, linear translations, and we include it as a testbed for this purpose. It is a generally easier dataset - this is clearly visible from spBLEU scores that our models achieve on it in Table~\ref{tab:results}.

\section{Minor Variations}
\label{app:minor}

\paragraph{Pretrained encoder and decoder} We tried using a pretrained encoder and decoder at initialization of our model and tested this for \texttt{hi-mr} and \texttt{es-ca}. For the former, the encoder and decoder were initialized with Hindi BERT \citep{joshi2022l3cube}, and for the latter, we used Spanish BERT \citep{CaneteCFP2020}. These pre-trained models are language-specific instances of BERT \citep{devlin2018bert}. Note that this means that we also used the pretrained tokenizers of these models, of sizes $52000$ and $31002$ for Hindi and Spanish respectively, that are only trained on the high-resource source languages; this leads to very poor tokenization in the target language. In general, this set of models take longer to converge due to their size, and show only minor differences in performance. Another related idea is to finetune \textsc{NLLB} or another multilingual MT model with an incorporated PGN; we did not try this given the lack of encouraging results from these experiments.

\paragraph{Single attention head} We also tried using only a single attention head to calculate $p_{copy}$ for target tokens, with the motivation that it was maybe better to nudge a single head to encode information about whether target token need to be copied, and leaving other heads to generate, as opposed to asking all heads to do both (which is the case when we average over heads). However, these models give almost identical results as in Table~\ref{tab:results}.

\paragraph{Smaller tokenizer} We hypothesize that using a smaller tokenizer size will force more splits per token, increasing the chance that common stems will be reflected in shared subwords. Accordingly, we tried a  tokenizer size of $8000$ for \texttt{hi-mr} and \texttt{es-ca} for the $15k$ and $60k$ settings; however, performance degrades slightly (about $-1$ spBLEU on average) for both \textsc{NMT} and \textsc{PGN} approaches without affecting the relative trend. 

This is not altogether surprising: reducing tokenizer size only increases the degree of splitting in words of a certain (lower) frequency range, rather than affecting the number of splits for all words uniformly. More importantly, while these hyperparameters are important to tune, statistical frequency-based tokenizers behave inherently differently from morphologically-inspired tokenizers, as discussed in Section~\ref{sec:disc}, and it is not easy or perhaps possible to achieve a good approximation of the latter by playing with the hyperparameters of the former.

\paragraph{Identical source and target}
Finally, we also trained a Hindi-Hindi model, to remove the effects of noisy translations and non-ideal tokenization of source and target token sequences as discussed in Sections~\ref{sec:data}~and~\ref{sec:disc}. In this setup, with $100\%$ overlap, the models achieve high test scores ($74$ spBLEU) and converge to near-zero usage of the copy mechanism. Clearly, the model still prefers to encode the identity relationship using a generate mechanism.

Note that this setup is fundamentally different from our other scenarios - when all tokens are copied, the model no longer needs the distinction between two distinct processes (generating and copying), and therefore does not really need to learn how to make this decision. However, it is still illustrative in demonstrating that model generalization mechanisms, even for highly simplified or trivial tasks, are often not intuitive or human-interpretable.

\section{Visualizations}
\label{app:vis}
\begin{figure*}
    \centering
    \begin{subfigure}{1\textwidth}
        \includegraphics[width=1\linewidth,scale=1]{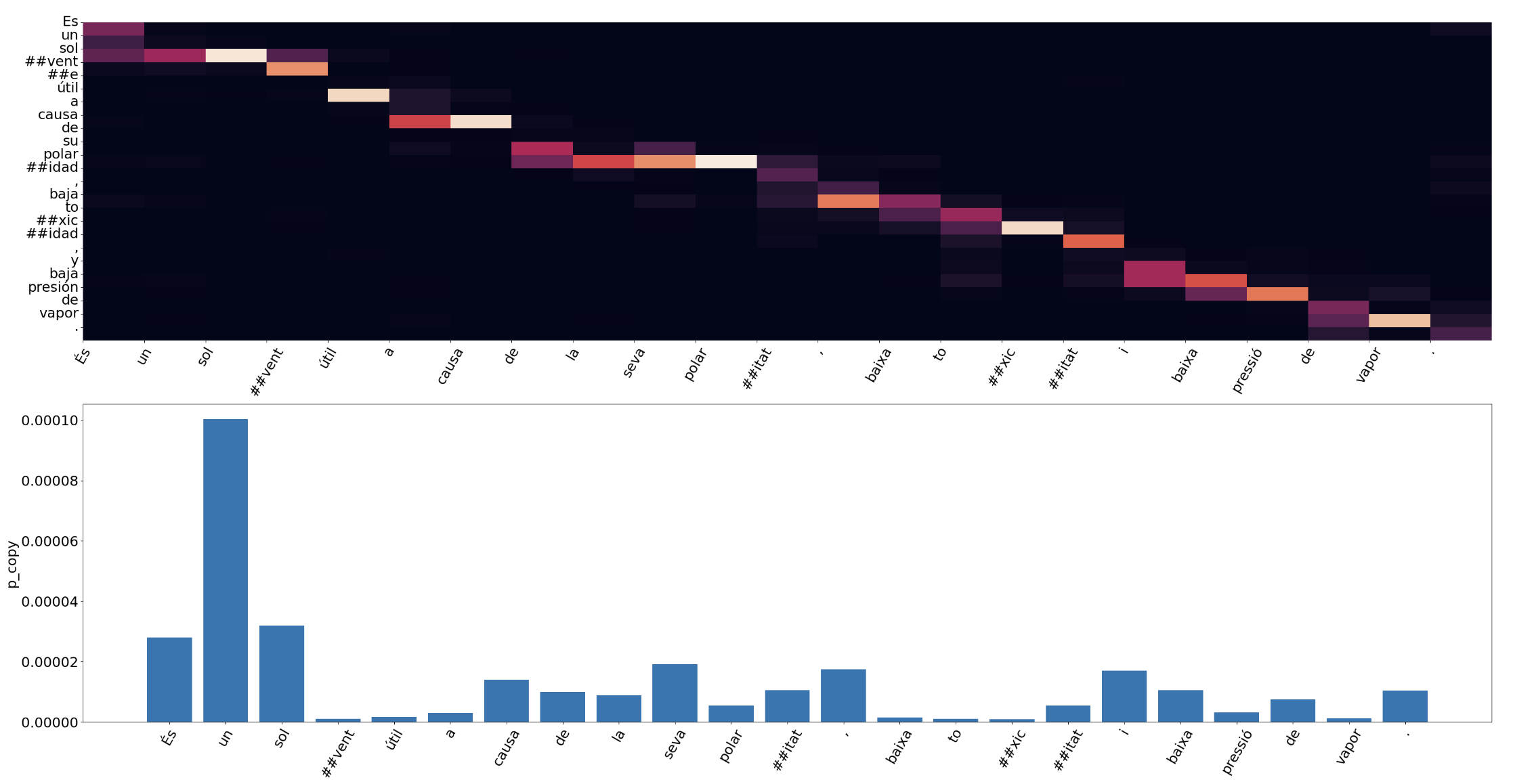}
        \caption{Epoch 10}
        \label{fig:esca_1}
    \end{subfigure}%

    \begin{subfigure}{1\textwidth}
        \includegraphics[width=1\linewidth,scale=1]{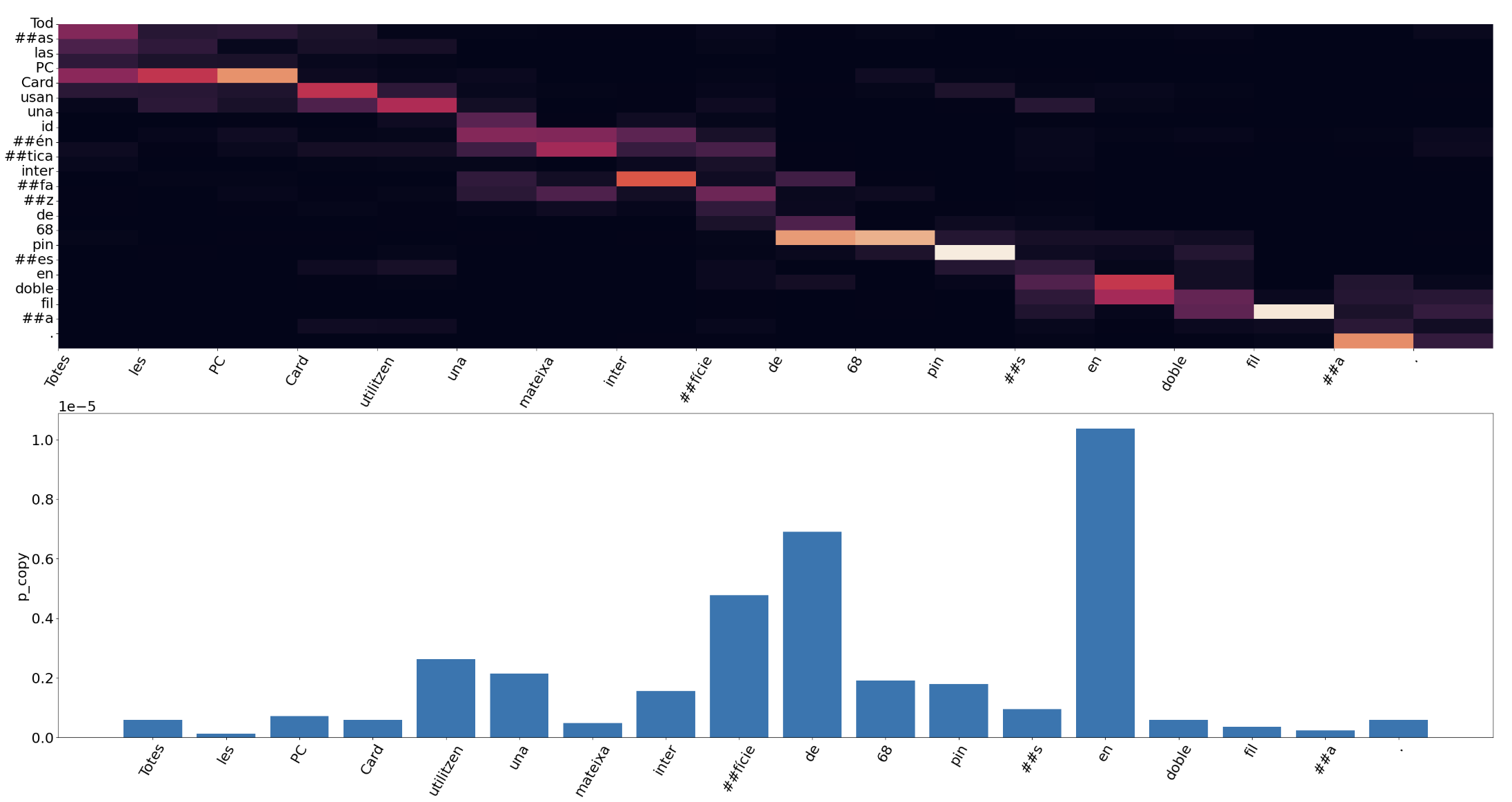}
        \caption{Epoch 30}
        \label{fig:esca_2}
    \end{subfigure}%
    \caption{Model's cross-attention distributions and $p_{copy}$ values for two sentence pairs for \texttt{es-ca(ep)}, $60k$ sentences}
    \label{fig:esca}
   \end{figure*}

\begin{figure*}
    \centering
    \begin{subfigure}{1\textwidth}
        \includegraphics[width=1\linewidth,scale=1]{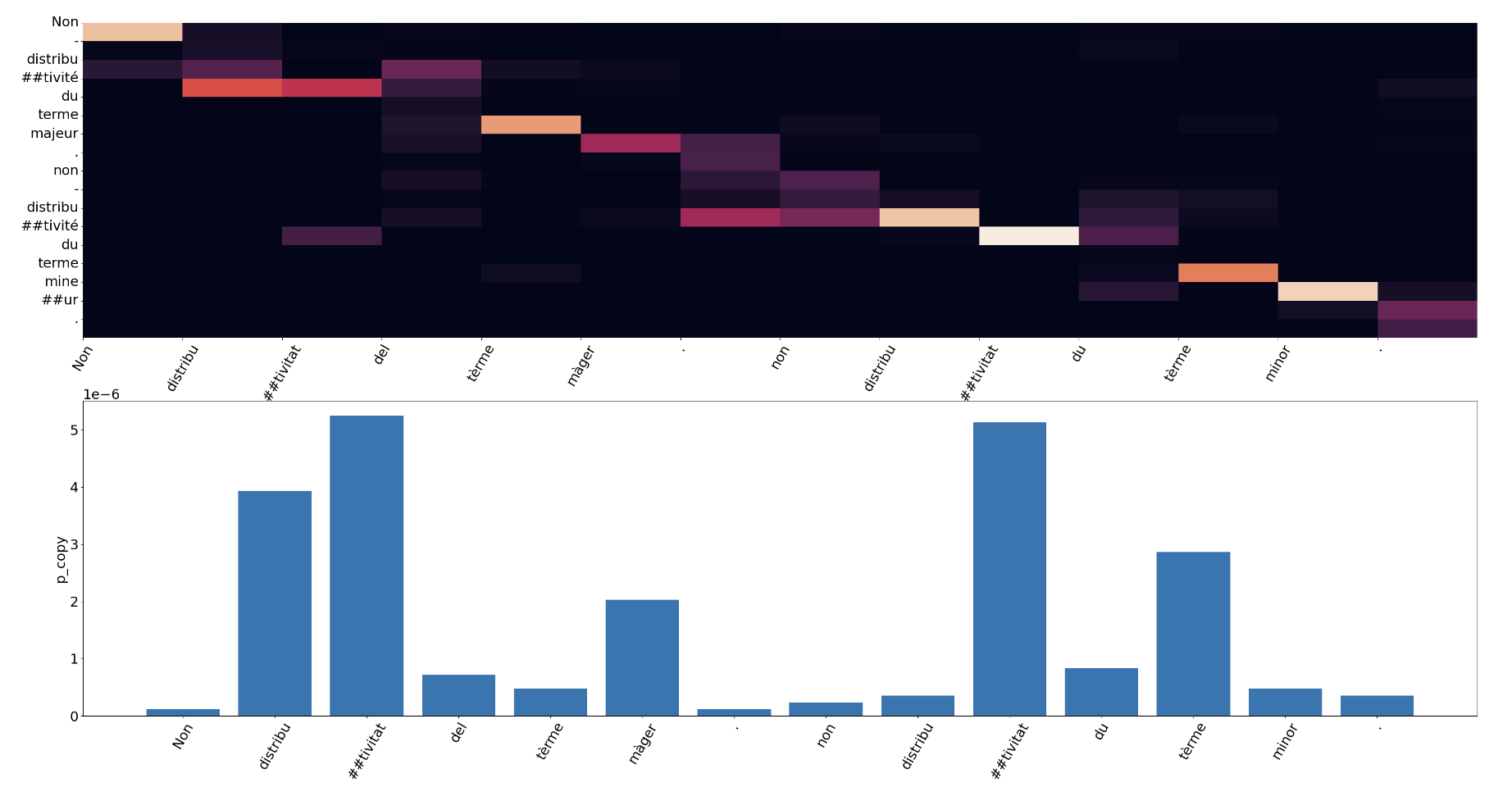}
        \caption{Epoch 10}
        \label{fig:froc_1}
    \end{subfigure}%

    \begin{subfigure}{1\textwidth}
        \includegraphics[width=1\linewidth,scale=1]{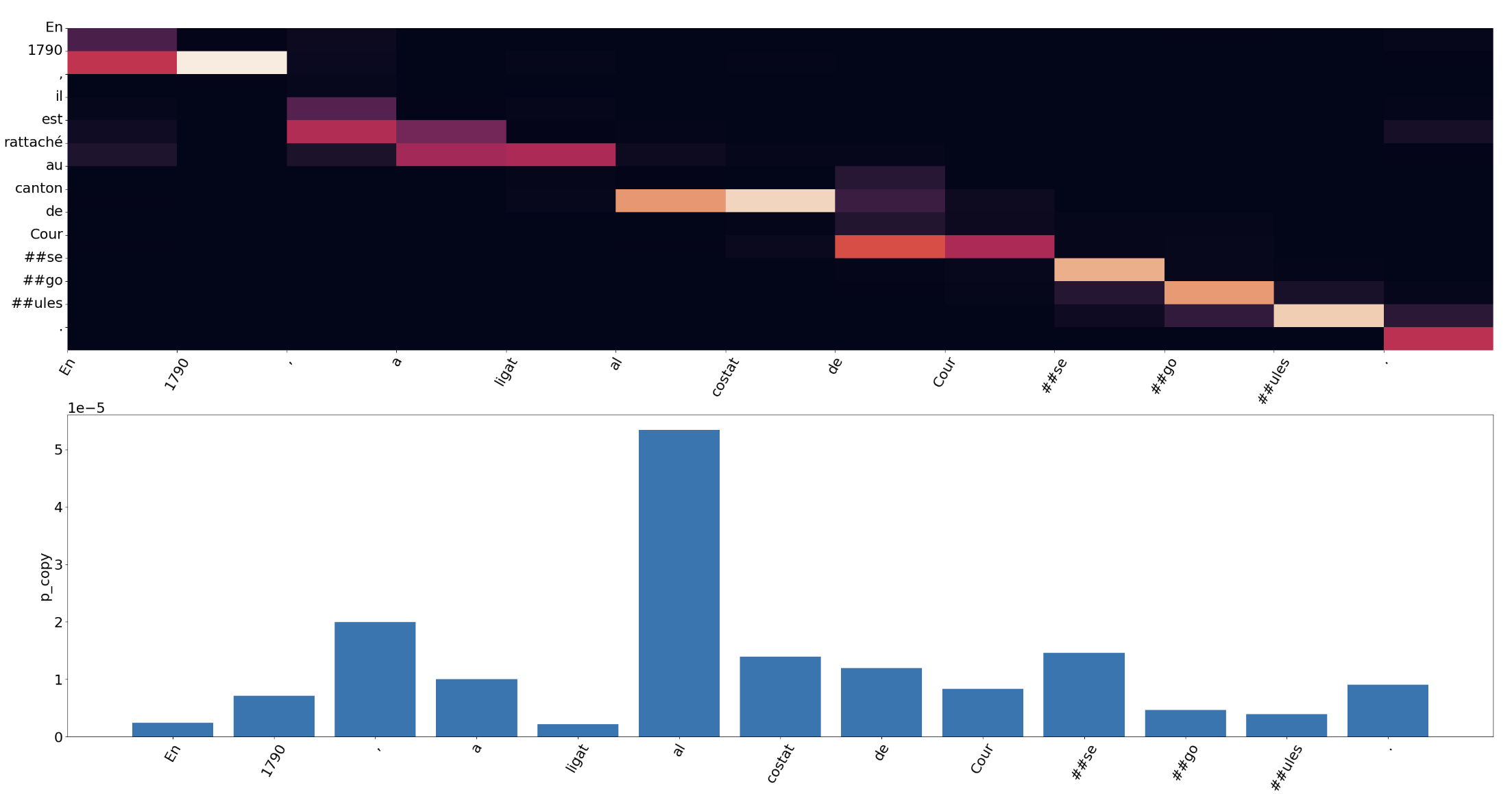}
        \caption{Epoch 30}
        \label{fig:froc_2}
    \end{subfigure}%
    \caption{Model's cross-attention distributions and $p_{copy}$ values for two sentence pairs for \texttt{fr-oc}, $60k$ sentences}
    \label{fig:froc}
   \end{figure*}

\begin{figure*}
    \centering
    \begin{subfigure}{1\textwidth}
        \includegraphics[width=1\linewidth,scale=1]{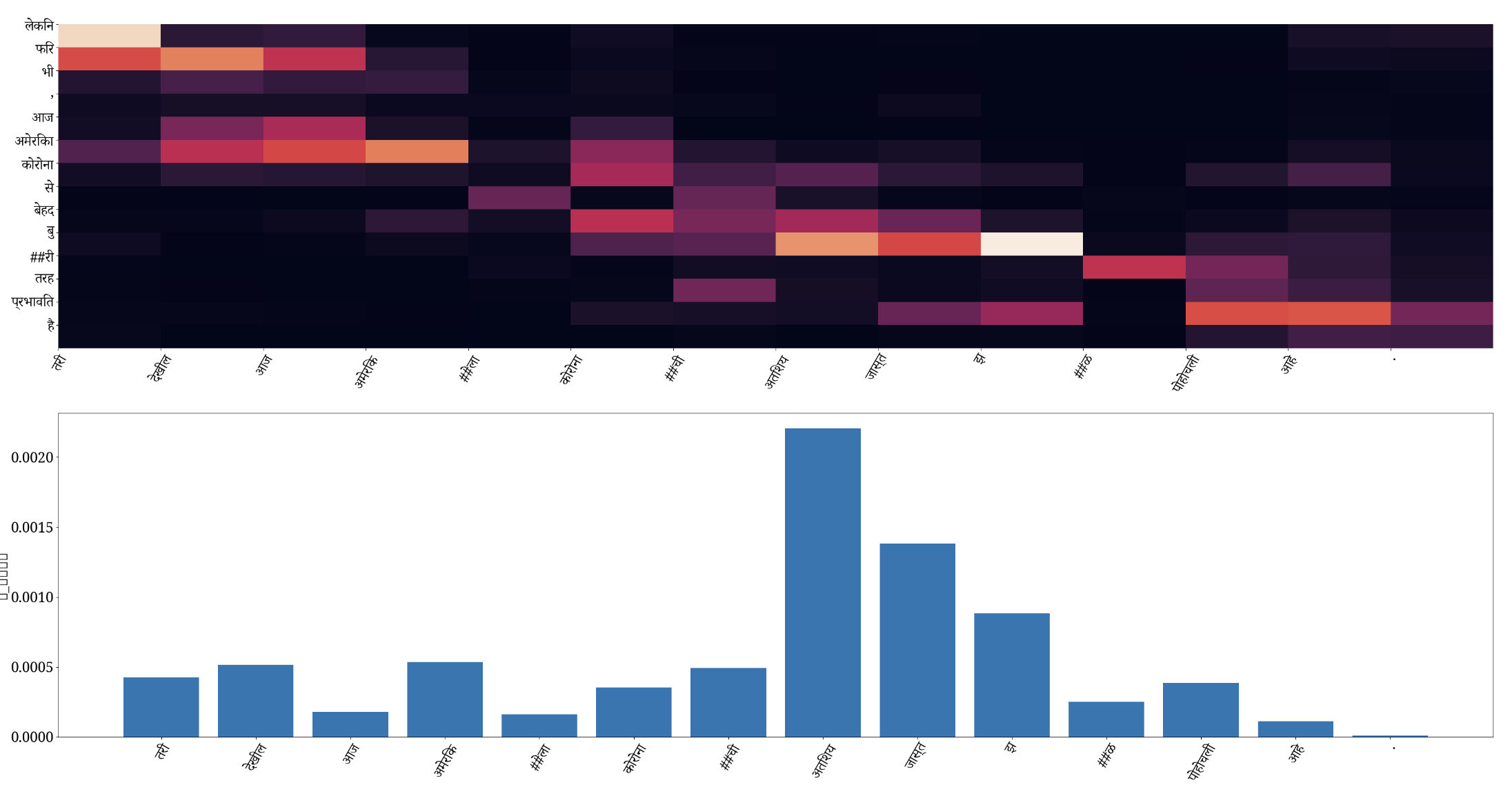}
        \caption{Epoch 10}
        \label{fig:himr_1}
    \end{subfigure}%

    \begin{subfigure}{1\textwidth}
        \includegraphics[width=1\linewidth,scale=1]{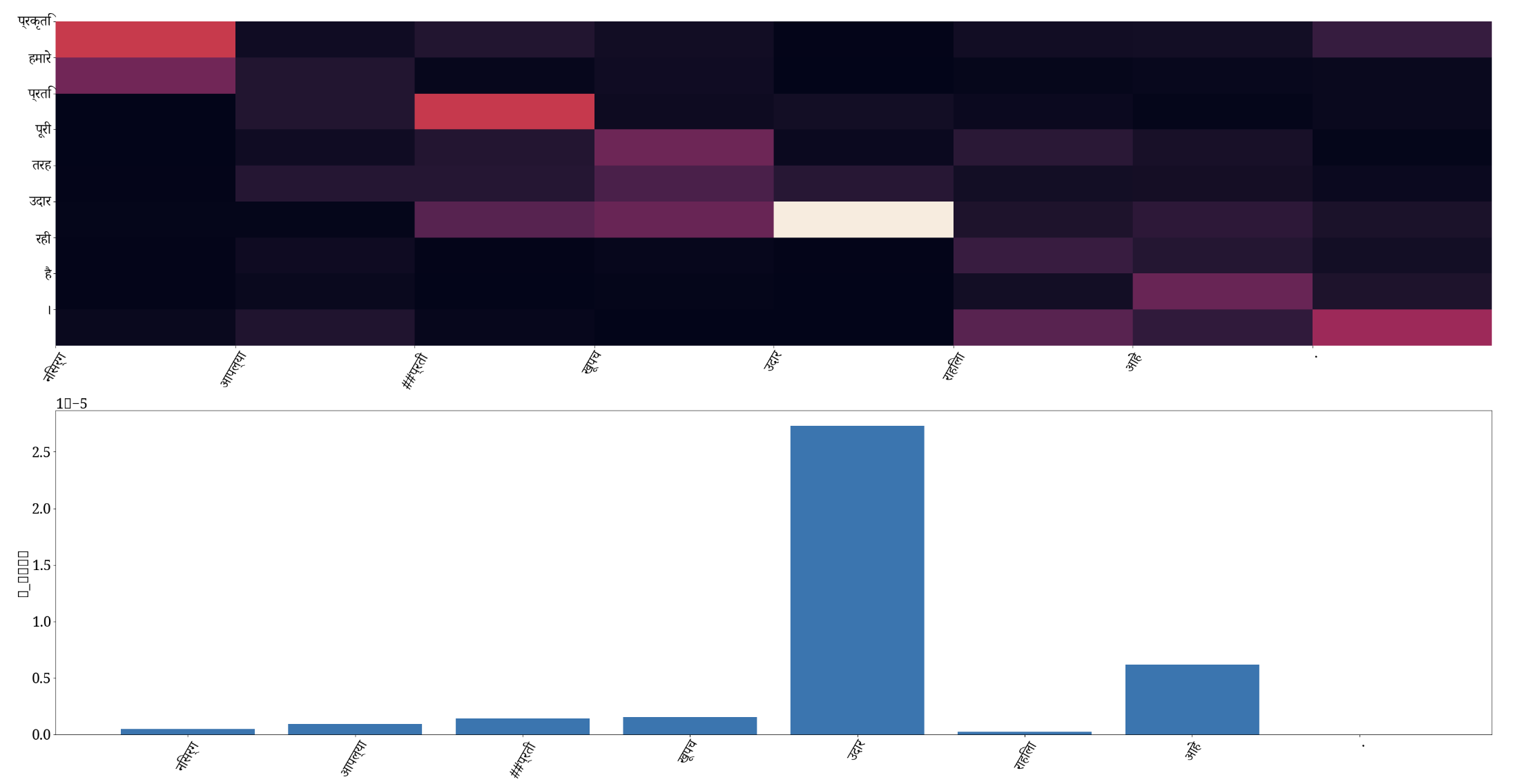}
        \caption{Epoch 25}
        \label{fig:himr_2}
    \end{subfigure}%
    \caption{Model's cross-attention distributions and $p_{copy}$ values for two sentence pairs for \texttt{hi-mr}, $60k$ sentences}
    \label{fig:himr}
   \end{figure*}

\begin{figure*}
    \centering
    \begin{subfigure}{1\textwidth}
        \includegraphics[width=1\linewidth,scale=1]{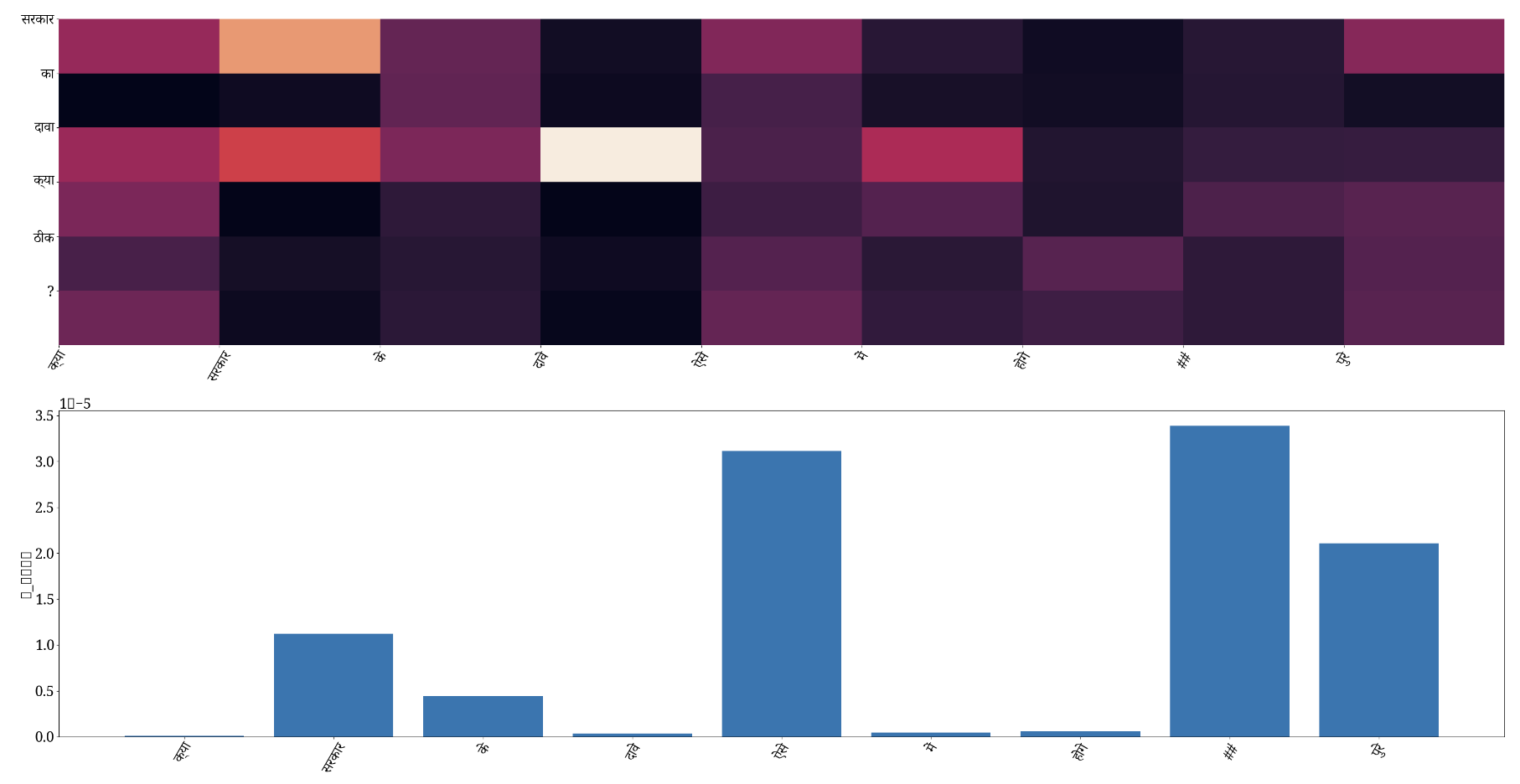}
        \caption{Epoch 10}
        \label{fig:hibh_1}
    \end{subfigure}%

    \begin{subfigure}{1\textwidth}
        \includegraphics[width=1\linewidth,scale=1]{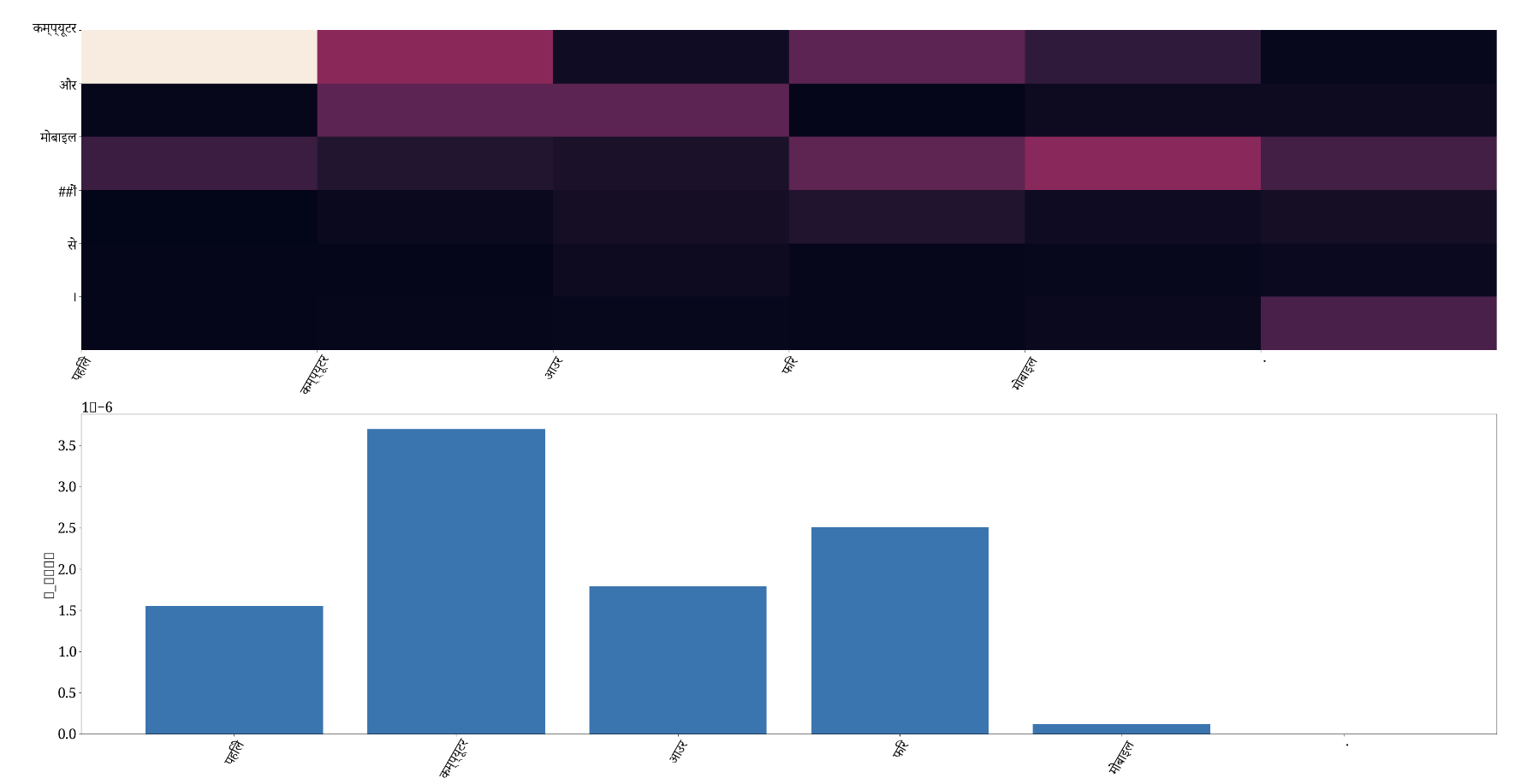}
        \caption{Epoch 30}
        \label{fig:hibh_2}
    \end{subfigure}%
    \caption{Model's cross-attention distributions and $p_{copy}$ values for two sentence pairs for \texttt{hi-bh}, $60k$ sentences}
    \label{fig:hibh}
   \end{figure*}

See Figures~\ref{fig:esca},~\ref{fig:froc},~\ref{fig:himr},~and~\ref{fig:hibh} for visualizations of the \textsc{PGN} model's cross-attention distributions and values of $p_{copy}$ per target token on randomly chosen source-target pairs using an early and late model training checkpoint. We observe that the model does use the copying mechanism as intended in many places, for common subwords (\texttt{udaar} in \texttt{hi-mr}, \texttt{un} in \texttt{es-ca}) as well as named entities (\texttt{Cour} in \texttt{fr-oc}), common borrowings (\texttt{computer} in \texttt{hi-bh}), numbers (\texttt{1970} in \texttt{fr-oc}) and punctuation. However, $p_{copy}$ values are also relatively high sometimes for other seemingly random target tokens, e.g. \texttt{canton-costat} in \texttt{fr-oc}. 

Note that the \texttt{hi-bh} source-target sentence pairs are not in fact translations of each other and exemplify the noise we discuss in Section~\ref{sec:data}~and Appendix~\ref{app:langs}. The cross-attention distributions for the \texttt{es-ca} and \texttt{fr-oc} models are in general much better defined and able to attend to appropriate tokens (in these example visualizations as well as others that we looked at); this is a consequence of the better quality of the data and models in these languages.

\end{document}